# A Scalable Chatbot Platform Leveraging Online Community Posts: A Proof-of-Concept Study

Sihyeon Jo
Electrical & Computer Eng.
Seoul National University
sihyeonjo@snu.ac.kr

Seungryong Yoo
Nuclear Engineering
Seoul National University
bastian95@snu.ac.kr

Sangwon Im
Aerospace Engineering
Seoul National University
born9507@snu.ac.kr

Seung Hee Yang
Interdisciplinary Program in
Cognitive Science
Seoul National University
sy2358@snu.ac.kr

Tong Zuo
Electronics & Information Eng.
North China Univ. of Technology
zuotong1996@hotmail.com

Hee-Eun Kim
Curatorial Team
Art Center Nabi
hien@nabi.or.kr

SangWook Han
Nabi Lab
Art Center Nabi
rewmlif@nabi.or.kr

Seong-Woo Kim
Graduate School of Engineering Practice
Seoul National University
snwoo@snu.ac.kr

Abstract

The development of natural language processing algorithms and the explosive growth of conversational data are encouraging researches on the human-computer conversation. Still, getting qualified conversational data on a large scale is difficult and expensive. In this paper, we verify the feasibility of constructing a data-driven chatbot with processed online community posts by using them as pseudo-conversational data. We argue that chatbots for various purposes can be built extensively through the pipeline exploiting the common structure of community posts. Our experiment demonstrates that chatbots created along the pipeline can yield the proper responses.

## 1. Introduction

Human-computer conversation is one of the most challenging AI problems. Although much effort has been made decades ago, research progress seems to have been slow due to the lack of the dialogue datasets. Only recently has the explosive growth of social media allowed the growth of utterance-response (or Q-R) pairs, however, collecting sufficient Q-R pairs is still intractable for many specific domains.

In this paper, we consider a simplified version of the human-computer conversation task, which is called **short text conversation (STC)**. Deriving the appropriate response corresponding to the user's query is a key part of the STC. Our research hypothesis is that implementing STC leveraging online community data is possible.

The contributions of this paper include (1) proposal of the pipeline for STC exploiting the structure of online community post-comments data which makes building chatbots coping with a variety of topics scalable, (2) empirical verification of effectiveness for the system with a simple response retrieval model.

## 2. Related Work

Previous studies have approached short text conversation task in a rule-based (Weizenbaum, 1966) and learning-based (Williams and Young, 2007) fashion. These approaches have less dependence on Q-R dataset, but are costly in that they require lots of hand-crafted engineering effort. An alternative approach was to utilize a knowledge base consisting of large amounts of Q-R pairs.

Table 1. The common structure of online community post and associated comments.

| Post | Title | |
|---|---|---|
| | Main Text | |
| Comments | Comment 1 | Likes/Dislikes |
| | Comment 2 | Likes/Dislikes |

The alternative methods for the STC task can be classified into two categories: **Generation-based STC** (e.g., Xing *et al.*, 2017) and **Retrieval-based STC** (e.g., Wu *et al.*, 2018). Generation-based method is to generate an appropriate response as a sequence of tokens, relying heavily on the Q-R dataset for training data. Retrieval-based method relatively has a strong point in fluency of yielded responses, but also has a drawback that collecting sufficient amounts of Q-R pairs is costly.

Any data-driven approaches to the STC require conversational dataset on a large scale. In this paper, we demonstrate online community data as a plentiful source that can easily be used as pseudo-conversational data so that chatbots having different types of character can be extensively built up. Our experiment suggests the potential of the platform for adaptable chatbot development.

## 3. Conversation on Online Community

Most online communities have a similar structure on posts. The general structure of online community posts is shown in Table 1. A post consists of title and main text, and related comments are listed with user's score represented as likes and dislikes.

We insist that online community user interaction, which is being generated at a tremendous speed and volume, can be used as a chatbot dataset with simple preprocessing. Online community posts have potential to be decent source for data-driven chatbot in terms of dealing with various topics and diverse groups of users which are possible to be reflected in the chatbots' tones and characters.

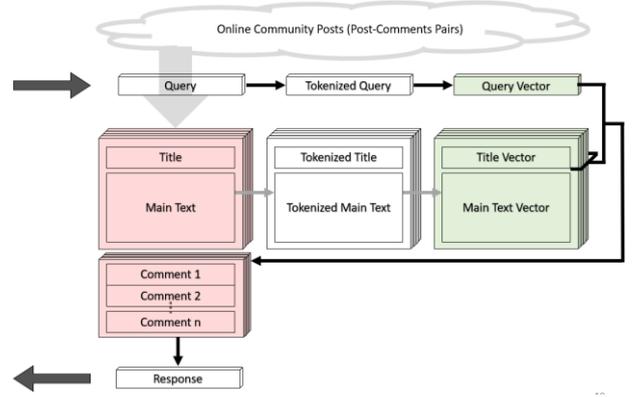

Figure 1 Retrieval-based STC system pipeline

## 4. Retrieval-based Short Text Conversation

### 4.1 Task Description

Short text conversation (STC) is defined as a task that derives a proper response for the user's query. In this paper, we approach the STC task in retrieval-based manner for quick verification on the feasibility of community dataset.

Formally, for a given query $q$, let $D$ be the given post-comments pair. The retrieval-based short text conversation system retrieves an appropriate response based on the following three stages.

- Step 1 (Retrieval), in which the system retrieves $I$ that is a subset of $D$ based on $q$.

$$I = Retrieve(q, D)$$

- Step 2 (Matching), in which the system selects candidate responses $C$ from $I$ by comparing matching scores calculated between $q$ and $I$.

$$C = Match(q, I)$$

- Step 3 (Ranking), in which the system ranks all response candidate $r$ in $C$ and chooses proper response $\hat{r}$:

$$\hat{r} \in \underset{r \in C}{argmax}\ Rank(q, r)$$

### 4.2 System Architecture

The overall system structure is illustrated in Figure 1. The gray arrows in Figure 1 work offline and the black arrows work online. After text embedding is completed, the system performs retrieval-based STC following the three steps explained in Section 4.1.

Table 2. An illustrative example of the system results.

| | |
|---|---|
| Query | 오늘 헤어졌어. |
| | *Broke up today.* |
| Responses | ㅋㅋㅋㅋㅋㅋㅋㅋㅋㅋㅋ |
| | *LOL* |
| | 저도 그런 사람이 있었어요. 왜 헤어지신 거예요? |
| | *I also had someone like him/her. Why did you break up?* |
| | 이건 거울 말도 들어봐야... |
| | *Let's look in the mirror...* |

### 4.3 Short Text Conversation Dataset

After crawling the post-comments pair from online communities, we excluded non-text-based posts such as photos and music, and also kept out explicit noises, for example advertisements, from the corpus. We then used KoNLPy (Park and Cho, 2014) to tokenize all the texts into parts of speech with normalization. During the process, postposition, punctuation marks, and emojis can be excluded to remove the characteristics of web writing up to designer's choice.

## 5. Results

Building the pipeline proposed in Section 4, we experiment by constructing a simple STC model using the dating counseling forum data for about 11 years (from December 15th, 2008 to October 1st, 2019). The total number of posts is 115,580 and only 105,206 posts are used as a corpus following the rule described in Section 4.3.

Using post-comments data from the dating counseling forum in SNULife is a proper choice to quickly verify that online community data can be utilized as pseudo-conversational data. Many posts deliver questions or concerns clearly, and most of the comments are relevant to the author's questions. These features of the data are advantageous when starting with the small amount of data to prove the research hypothesis.

Doc2vec (Le *et al.*, 2014) and TF-IDF are used for text embedding. The dimension of vectors with doc2vec is 2,000 for both title and main text, while the size of vectors embedded through TF-IDF is 24,795 for title corpus and 105,878 for main text corpus which represent the size of vocabulary in each corpus.

### 5.1 Candidate Responses Retrieval and Ranking

The retrieval and matching steps (described in Section 4.1) can be grouped into candidate responses retrieval process. In the retrieval process, we retrieve 200 posts ($I$ in Section 4.1) based on cosine similarity calculated with doc2vec vectors between the query and entire post titles. We then select five posts among 200 with cosine similarity between query and subset $I$ in TF-IDF embedding.

Finally, in the ranking process, at most two popular comments (measured by *Likes-Dislikes*) for each post are picked without embedding of comments. Although there are various standards to rank the final response, we decided to choose one among candidate responses randomly to increase user's satisfaction by avoiding the same answer to the same query.

### 5.2 Lessons Learned

Based on the proposed pipeline leveraging online community posts as a data source for building up chatbots, we conduct a quick attestation of the hypothesis. Our demonstration shows that the theory works to a certain extent as presented in Table 2. In particular, the atmosphere of the community is reflected in the chatbot's tone and mood.

The advantage of not using the information in the query directly when ranking the candidate responses is to make the most of the semantic relevance information of the post-comments data itself. However, query information is needed to deal with comments including external knowledge such as community memes or practices. Improved ranking module of the system to be addressed in our future work.

As more data can be collected with the constructed pipeline, we can develop more sophisticated deep learning models using the data. Our goal is to make a breakthrough from STC to multiple round conversation task in an end-to-end manner. Learning a distribution of proper responses with a nonlinear model such as the neural network and evaluating the model with quantitative performance measures are our research direction.

## 6. Conclusion

In this paper, we verified the possibility of constructing a data-driven chatbot by processing posts from the online community and using it as real-world conversation data.

We evaluated our hypothesis on the real-world online community dataset and chatbot scenarios, and obtained promising results. Leveraging the data collected from the built pipeline, multiple rounds of conversation to be addressed in our next research.

## Acknowledgement

This research was supported by the National Research Foundation of Korea (NRF) funded by the Ministry of Science and ICT (2018R1C1B5086557, 2019027648).